\documentclass[final,5p,times,twocolumn]{elsarticle}

\usepackage{hyperref}
\usepackage{lineno}
\usepackage{comment}
\usepackage{amsmath}
\usepackage{float}
\usepackage{graphicx}
\usepackage{svg}
\usepackage{pdfpages} 
\modulolinenumbers[5]
\usepackage{pgfplots}
\usepackage{amssymb}
\usepackage{multirow}
\usepackage{ gensymb }
\usepackage{multicol}
\usepackage{makecell}
\usepackage{caption}
\journal{arXiv}

\begin{document}
\begin{frontmatter}

\title{ \huge SmaAt-UNet: Precipitation Nowcasting using a Small Attention-UNet Architecture}

\author{Kevin Trebing}
\author{Tomasz Stańczyk}
\author{Siamak Mehrkanoon\corref{cor1}}

\cortext[cor1]{Corresponding author}

\address{Department of Data Science and Knowledge Engineering, Maastricht University, The Netherlands}

\begin{abstract}
Weather forecasting is dominated by numerical weather prediction that tries to model accurately the physical properties of the atmosphere. A downside of numerical weather prediction is that it is lacking the ability for short-term forecasts using the latest available information. By using a data-driven neural network approach we show that it is possible to produce an accurate precipitation nowcast. To this end, we propose \textit{SmaAt-UNet}, an efficient convolutional neural networks-based on the well known UNet architecture equipped with attention modules and depthwise-separable convolutions. We evaluate our approaches on a real-life datasets using precipitation maps from the region of the Netherlands and binary images of cloud coverage of France. The experimental results show that in terms of prediction performance, the proposed model is comparable to other examined models while only using a quarter of the trainable parameters.
\end{abstract}

\end{frontmatter}

\section{Introduction}
\label{sec:introduction}

Computational weather forecasting is an ubiquitous feature of modern, industrialized societies and is used for planning, organization and management of a wide range of both personal and economic aspects of life. To date, the primary method for weather forecasts is numerical weather prediction (NWP). NWP relies on mathematical models that take into account different physical properties of the atmosphere such as air velocity, pressure and temperature. The NWP-based models can generate accurate weather predictions of several hours to days into the future. However, they involve solving highly complex mathematical models which are computationally expensive and require enormous computing power and thus usually are performed on expensive supercomputers \cite{soman2010review}. 

Due to their high computational and time requirements, NWP models are less suitable for short-term forecasts ranging from minutes to up to 6 hours, also referred to as nowcasting \cite{hering2004nowcasting}. Nowcasting models are able to use the latest available observational weather data to create their predictions, making them more responsive compared to the NWP models \cite{dwdnowcasting}. This responsiveness is critical to increase the accuracy of predictions for dynamic and rapidly changing environments such as the atmosphere. Nowcasts have therefore become important tools to complement NWP approaches, especially in the context of meteorologically unstable conditions typical for severe weather hazards such as thunderstorms and heavy rainfall \cite{dwdnowcasting}. As highlighted by a status report to the American Meteorological Society, nowcasting thunderstorms finds pertinent applications across a variety of fields such as in aviation, the construction industry, power utilities and ground transportation \cite{wilson1998nowcasting}. Nowcasting was also used in the 2008 Olympic games in Beijing to ensure the safety of the athletes \cite{wilson2010nowcasting}. Not least, weather nowcasts can also be useful for planning ordinary activities of everyday life. 

Recent advances in artificial neural network architectures (ANNs) have enabled data-driven based models to bridge the present gap for short-term forecasting \cite{xingjian2015convolutional, wang2017predrnn, tran2019multi, sonderby2020metnet}. The key difference between NWP and a ANNs is that the former is a model-driven and the latter a data-driven approach. Unlike the model-driven approaches, data-driven models do not base their prediction on the calculations of the underlying physics of the atmosphere. Instead, they analyze and learn from historical weather data such as past wind speed and precipitation maps to predict the future. 

In this paper, we introduce a novel artificial neural network based model to predict precipitation on a high-reso\-lution grid 30 minutes into the future. The input data for our model consists of precipitation maps, i.e. cartographic radar images showing the accumulated rainfall over a period of time. In addition, the applicability of the proposed model is also shown on cloud cover nowcasting task.

In previous studies, convolutional neural networks have been described as an effective approach to process image data. Convolutions are kernel-based operations that slide over the image which enables the model to capture local invariant features in a more efficient manner than other feedforward approaches \cite{krizhevsky2012imagenet}. They have been successfully applied in various fields including not only the processing of images but also of other types of signals. For instance, the authors of \cite{anderson2018bottom} used a CNN-based model to create captions for an input image while \cite{he2016deep} employed a CNN for object detection in images. The authors in \cite{mehrkanoon2019deep} introduced a 3-dimensional CNN-based model to predict the wind speed in different cities in Denmark. In another study, a CNN-based architecture is applied on signals from a smartphone's accelerometer to classify a user's transportation mode \cite{liang2019deep}. The authors in \cite{trebing2020wind} introduced multidimensional convolutional neural networks for wind speed forecasting.

Given the usefulness of CNNs for tasks involving image input, they offer a promising solution for the purpose of precipitation nowcasting. In this paper, we propose Small Attention-UNet (SmaAt-UNet) model. It uses the UNet architecture \cite{ronneberger2015u} as core model and is equipped with attention modules and depthwise-separable convolutions (DSC). (see section \ref{sec:methods} for more details).

The advantage  of our model is that we are able to reduce the model parameter size to a quarter of the original UNet implementation while maintaining a comparable performance to the original UNet architecture. 

This reduction in model size opens up the possibility to the use of precipitation models on small computation units such as smartphones, similar to \cite{howard2017mobilenets}. This could enable the use of personalized and up-to-date precipitation forecasts by creating a forecast on user request with the latest available data within seconds. Furthermore, a model size reduction with similar performance than bigger models is crucial for creating efficient architectures that require less training and computational power. 

This paper is organized as follows. A brief overview of related research on weather forecasting using machine learning architectures is presented in section \ref{sec:related_work}. In section \ref{sec:methods}, we describe the proposed SmaAt-UNet architecture and other models for precipitation as well as cloud cover nowcasting.
Section \ref{sec:experiments} describes the conducted experiments and the obtained results. A discussion of the results is given in section \ref{sec:discussion}. Lastly, we end with some conclusive remarks in section \ref{sec:conclusion}.

\section{Related Work}
\label{sec:related_work}
A common approach to precipitation nowcasting based on deep learning uses neural networks that have some kind of memory such as a Long-short term memory (LSTM) \cite{hochreiter1997long}. 
In standard feedforward models, the input is passed on in a straight forward fashion from one timestep to the next. In contrast, LSTMs are, broadly speaking, networks that enable the input signal to remain in the network's state for multiple time steps enabling the network to remember past inputs. This is especially useful for time-series predictions because past inputs can have valuable information about trends which, in turn, can be useful for predicting future values. 

The authors in \cite{xingjian2015convolutional} created a convolutional LSTM that captures spatiotemporal correlations better than other approaches in a time-series task for images. Extending on this, the authors of \cite{wang2017predrnn} created a spatiotemporal-LSTM that increases the amount of memory connections inside the network which aims at enabling an efficient flow of spatial information. The memory function and memory flow of this model were optimized in another implementation that added stacked memory modules \cite{wang2019memory}. 

Another approach for precipitation nowcasting has been described in  \cite{agrawal2019machine}. They proposed a network structure that is based on a well-known encoder-decoder architecture called UNet \cite{ronneberger2015u}. Unlike LSTMs, UNet has no explicit modeling of memory. It takes an input image (or multiple concatenated images) and outputs a single classification map. The implementation of \cite{agrawal2019machine} aimed at classifying four different rain intensities ($<0.1mm/h$, $<1.0mm/h$, $<2.5mm/h$, $>2.5mm/h$) one hour into the future. To this end, multiple precipitation maps (of the past hour) are concatenated and used as input to the UNet architecture. In a similar study in \cite{sonderby2020metnet}, as opposed to the model described in \cite{agrawal2019machine} the authors classified 512 classes instead of just four, thereby resulting in a much finer resolution of rain intensities. This is similar to our approach; however, rather than predicting classes, our model predicts exact rain intensities. 

A common baseline in precipitation nowcasting is the persistence method. The persistence model uses the last input image of a sequence as the prediction image. This is based on the assumption that the weather will not change significantly from time point $t$ to $t+1$. Especially in nowcasting this baseline is not trivial to be outperformed because the time differences between images are so short (e.g., 2 or 5 minutes) that often weather conditions remain the same \cite{soman2010review}.

Recently, it was shown that attention in CNNs can be a very useful tool to enhance performance for an underlying task \cite{hu2018squeeze, bello2019attention, jaderberg2015spatial, oktay2018attention, zhang2018shufflenet}. Attention is a mechanism that amplifies wanted signals and suppresses unwanted ones. This directs the network to pay more attention to features important for the task at hand. In our proposed model, we employ convolutional block attention modules (CBAMs) that take the input image and apply attention in sequence to the channels and then to the spatial dimensions \cite{woo2018cbam}. The result of a CBAM is a weighted feature map that takes into account the channels and also the spatial region of the input image. 

To the best of our knowledge, it is the first time to include CBAM mechanism within a UNet-based architecture.

In another application of attention, authors of \cite{oktay2018attention} added attention gates to a UNet architecture for a medical segmentation task. They found that their enhanced model achieved better results than the original UNet implementation by \cite{ronneberger2015u}.

Having fewer parameters in a network reduces the chance of possible overfitting, because the model is simpler and can't adapt too closely to the training set's distribution. A possible downside to this simplification is that the model may be too simple to learn the desired task.
In order to reduce the number of parameters without sacrificing a lot of performance, depthwise-separable convolutions (DSC) are used in many recent architectures \cite{lawhern2018eegnet,chollet2017xception, guo2019depthwise, howard2017mobilenets, gadosey2020sd}. DSCs split up the regular convolutional operation into two separate operations: a depthwise convolution followed by a pointwise convolution. This results in fewer mathematical operations and also fewer parameters compared to non-separated convolutions. The authors in \cite{gadosey2020sd} created a UNet with DSCs instead of regular convolutions and their model has eight times less parameters than the original UNet implementation. They show that their model is able to have a similar performance as UNet on medical segmentation tasks \cite{gadosey2020sd}.

\section{Methods}
\label{sec:methods}

\subsection{Proposed SmaAt-UNet}
\begin{figure*}
    \centering
    \includegraphics[width=0.8\linewidth]{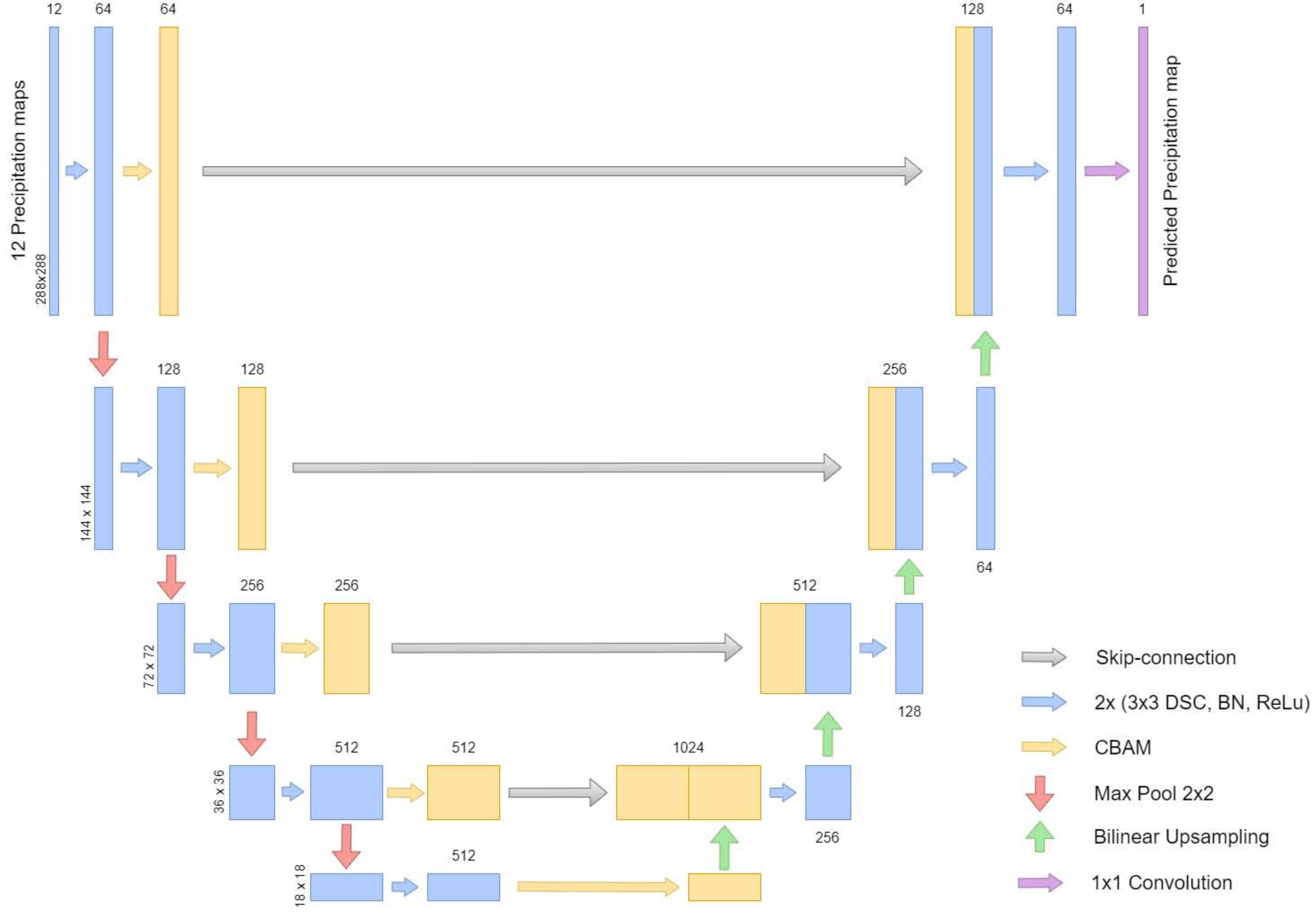}
    \caption{An example of an input fed through our proposed SmaAt-UNet (best viewed in color). Each bar represents a multi-channel feature map. The numbers above each bar display the amount of channels; the vertical numbers on the left side correspond to the x-y-size.} 
    \label{fig:model}
\end{figure*}

The model that we propose here builds upon and extends the UNet architecture \cite{ronneberger2015u}. As shown in Figure~\ref{fig:model}, the UNet architecture consists of an encoder-decoder structure resulting in a U-shape. The en\-coder-part (corresponding to the left half of Fig.~\ref{fig:model}) applies max-pooling (red arrows) and a double convolution (blue arrows) which halves the image size and doubles the number of feature maps, respectively. The encoders are subsequently followed by the same amount of decoders (corresponding to the right half of Fig.~\ref{fig:model}).
Following the original implementation of UNet, here we also use four encoder-decoder modules. 

A decoder consists of three parts: a bilinear upsampling operation (green arrows) to double the feature map size, a concatenation of the resulting feature maps with the previous encoder's output via skip-connections (grey arrows), and lastly a double convolution to half the number of feature maps. The skip-connections enable the model to use multiple scales of the input to generate the output. Finally, the last layer in our model is a $1 \times 1$ convolution (purple arrow) which outputs a single feature map representing the value predicted by the network.

The advantage of using different scales is that they can capture differently sized objects in the input which can be important for some tasks. Typically, UNets are applied to classification or segmentation tasks in which the network is trained to predict a class for each pixel. However, we applied it to a time series prediction task in which the network has to predict an exact value for each pixel.

Our novel Small Attention-UNet (SmaAt-UNet) model makes two modifications in the original UNet architecture. Firstly, we propose to add the CBAM attention mechanism to the encoder part. Secondly, we transform the regular convolutional operations to depthwise-separable convolutions. 

As described in section \ref{sec:related_work}, using attention in a CNN facilitates the network to focus on specific parts of the input. For our model, we use convolutional block attention modules for the purpose of identifying important features across channels and spatial regions of the image \cite{woo2018cbam}. 
In CBAMs, the attention mechanism is applied first across the channels of the image and subsequently to the spatial dimension. 

The CBAMs are placed after the first double convolution and every encoder to amplify important features and suppress unimportant ones on the respective image scale (yellow arrows in Fig.~\ref{fig:model}). Importantly, the input to the encoders is the convoluted and downsampled image from the previous encoder and not the image with the attention mechanism applied. This way, the original image features are preserved until the last encoder. The attention modules only feed into the corresponding upsampling part of the network to which they are connected through the skip-connections.

Following the lines of \cite{lawhern2018eegnet, howard2017mobilenets}, we used depth\-wise-separ\-able convolutions in our model in order to reduce the number of parameters. In particular, we substitute all convolutions of the original UNet model with depthwise-separable convolutions. However, in the convolutional block attention modules we still apply regular convolutions. 

\subsection{Other models}
\label{sec:other_models}
For comparison, we also trained other UNet architectures that have either none or only one of the two modifications that we proposed. This results in a total of four models being compared in this study, i.e. the original UNet, UNet with CBAM, UNet with DSCs, and our proposed model. Table~\ref{tab:parameters} shows a comparison of the models' parameters. When looking at the standard UNet architecture and our proposed modified UNet architecture it can be seen that the latter has significantly fewer parameters, i.e. $\approx$17m compared to $\approx$4m. 
In our PyTorch implementation\footnote{available at \url{https://github.com/HansBambel/SmaAt-UNet}} we use DSCs with two kernels-per-layer. 

\begin{table}[]
    \centering
    \caption{Number of parameters of the compared models.}
    \label{tab:parameters}
    \begin{tabular}{|c|c|}
    \hline
         Model & Parameters \\
         \hline
         UNet & 17,272,577 \\
         UNet with CBAM & 17,428,781 \\
         UNet with DSC & 3,955,185 \\
         SmaAt-UNet & 4,111,389 \\
         \hline
    \end{tabular}
\end{table}

\subsection{Training}
All four previously described models were trained for a maximum of 200 epochs. We employed an early stopping criterion which stopped the training process when the validation loss did not increase in the last 15 epochs. The early stopping criterion was met in all training iterations so that the maximum of 200 epochs was never reached. Additionally, we used a learning rate scheduler that reduced the learning rate to a tenth of the previous learning rate when the validation loss did not increase for four epochs. The initial learning rate was set to 0.001 and we used the Adam optimizer \cite{kingma2014adam} with default values. The training was done on a single NVidia 2070 Super with 8Gb of VRAM.

\subsection{Model evaluation}
The loss function used in this study is the mean squared error (MSE) between the output images and the ground truth images. The MSE is calculated as follows:
\begin{equation}
    MSE = \frac{\sum^n_{i=1} (y_i - \hat{y}_i)^2}{n}
\end{equation}
where $n$ is the number of samples, $y_i$ is the value of the ground truth and $\hat{y}_i$ is the value of the prediction.

In addition to the MSE, we calculate different scores for performance evaluation, such as Precision, Recall (probability of detection), Accuracy and F1-score, critical success index (CSI), false alarm rate (FAR) and Heidke Skill Score (HSS). In case of the precipitation map dataset, these scores are calculated for rainfall bigger than a threshold of $0.5mm/h$. To do this, we convert each pixel of the predicted output and target images to a boolean mask using this threshold. In case of the cloud cover dataset, the data is already binarized\footnote{Both datasets are described in section~\ref{sec:experiments}.}. From this, one can calculate the true positives (TP) (prediction=1, target=1), false positives (FP) (prediction=1, target=0), true negatives (TN) (prediction=0, target=0) and false negatives (FN) (prediction=0, target=1). Subsequently the CSI, FAR and HSS metrics can be computed as follows:
\begin{equation}
\textrm{CSI}=\frac{TP}{TP+FN+FP},
\end{equation} 
,
\begin{equation}
\textrm{FAR}=\frac{FP}{TP+FP},
\end{equation}
and
\begin{equation}
    \textrm{HSS}=\frac{TP\times TN - FN\times FP}{(TP+FN)(FN+TN)+(TP+FP)(FP+TN)}.
\end{equation}
The threshold of $0.5mm/h$ (the first dataset) was chosen in line with the works by \cite{xingjian2015convolutional, shi2017deep} and it differentiates between rain and no rain.

\section{Experiments}
\label{sec:experiments}
\subsection{Precipitation map dataset}
We used a precipitation data from the Royal Netherlands Meteorological Institute (Koninklijk Nederlands Meteorologisch Instituut, KNMI) as the first dataset to train and compare our models. It contains rain maps in 5-minute intervals from the last four years (2016-2019) of the region of the Netherlands and the neighboring countries. In total, the dataset comprises about 420,000 rain maps. The data is generated by two C-band Doppler weather radar stations situated in De Bilt (52.10$^{\circ}$N, 5.18$^{\circ}$E, 44 m MSL) and Den Helder (52.96$^{\circ}$N, 4.79$^{\circ}$E, 51 m MSL), the Netherlands. To acquire a rain map, the two radars perform four azimuthal scans of 360$^{\circ}$ around a vertical axis beam elevation angles of 0.3$^{\circ}$, 1.1$^{\circ}$, 2.0$^{\circ}$, and 3.0$^{\circ}$. Additional parameters of the radars can be found in Table~\ref{tab:radars}. Furthermore, the rain maps are rain-gauge adjusted with more details being described in \cite{overeem2009derivation}. We split up the dataset into a training set (years 2016-2018) and a testing set (year 2019). Additionally, for every training iteration, a validation set was created by randomly selecting 10\% of the training set.

\begin{table}[]
    \centering
    \caption{Parameters of the two radars}
    \label{tab:radars}
    \begin{tabular}{|c|c|c|}
    \hline
         Characteristics & De Bilt & Den Helder  \\
         \hline
         Wavelength (cm) & 5.293 & 5.163\\
         Pulse repetition frequency (Hz) & 250 & 250 \\
         Peak power (kW) & 268 & 264 \\
         Pulse duration ($\mu$s) & 2.02 & 2.04\\
         3-db beamwidth ($^{\circ}$) & 1 & 1\\
         Antenna rotation speed ($^{\circ}$ $s^{-1}$) & 18 & 18 \\
         No. of samples per range ($km^{-1}$) & 4 & 4\\
    \hline
    \end{tabular}
    
\end{table}

The raw rain maps have a dimension of $765 \times 700$ and one pixel corresponds to the accumulated rainfall in the last five minutes on one square kilometer. The amount of rainfall is noted as an integer value in the unit of a hundredth of millimeter. For instance, a value of 12 means there was 0.12mm of rainfall in the last five minutes. 

\begin{figure*}
    \centering
    \includegraphics[width=.85\linewidth]{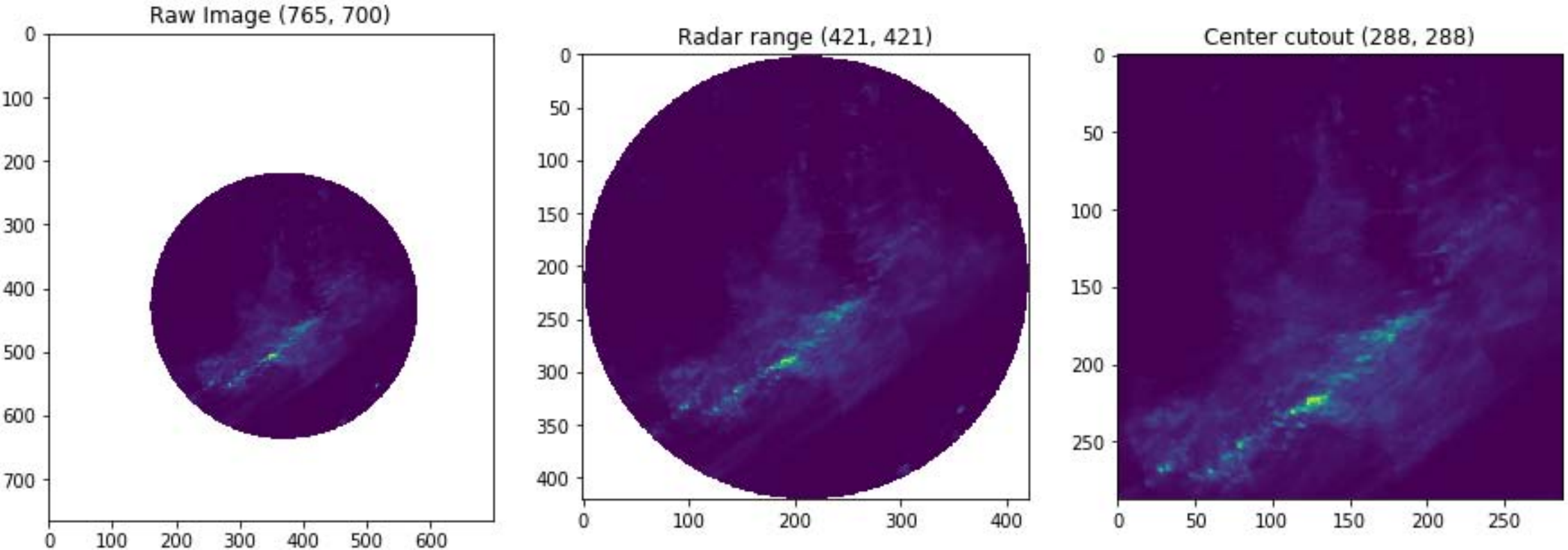}
    \caption{Transformations applied on the raw data (left): Cutout of max range of radars (middle) followed by center crop of 288 pixels (right).}
    \label{fig:cutout}
\end{figure*}
As a data preparation step, we divided the values of both the training and testing set by the highest occurring value in the training set to normalize the data. Furthermore, we cropped the image and only used a subset of the original precipitation map (Fig.~\ref{fig:cutout}). This was done due the fact that many pixels of the raw image have no-data values because the raw image is larger than the maximum range of the radar (see the white margin in the left panel of Fig.~\ref{fig:cutout}). The area within the range of the radar has a circular shape and a diameter of $421$ pixels corresponding to $421$ kilometers. When cropping the image in a way that preserves the entire radar image there are still many pixels with no-data values (white corners in the middle panel of Fig.~\ref{fig:cutout}). Since training a neural network with no-data values is more difficult, we therefore applied an additional center crop of 288 pixels (right panel of Fig.~\ref{fig:cutout}).

The input for the models is a sequence of 12 precipitation maps which are stacked along the channel dimension. This corresponds to one hour of past weather observations (12 x 5min). The output is the precipitation map 30 minutes later than the last input image. Therefore, the task for the network is to predict exact rainfall intensities for every pixel of the $288 \times 288$ rain map 30 minutes into the future.

The dataset contains many rain maps with very little to no rain. Therefore, in order to avoid biasing the network towards predicting zero values we created two additional data\-sets whose target images have a minimum amount of rainy pixels.
One of the two datasets has samples with at least $20\%$ of rainy pixels in the target images and the other one with at least $50\%$; we will call them NL-20 and NL-50, respectively. The number of samples in these two datasets is necessarily significantly smaller than the original dataset, but they also more appropriately apply to the use-case of the model, i.e. predicting rain. A comparison of different sample sizes of the three datasets can be found in Table~\ref{tab:datasets}. The data sets can be obtained by submitting a request to the authors.
\begin{table}[]
    \centering
    \caption{Comparison of the dataset sizes. The original dataset has a lot of images with little to no rain.}
    \label{tab:datasets}
    \resizebox{\columnwidth}{!}{%
    \begin{tabular}{|c|c|c|c|c|}
    \hline
         Name & Required rain pixels & Train & Test & Subset \\
         \hline
         NL-Full & 0\% (original) & 314940 & 105003 & 100\% \\
         NL-20 & 20\% & 31674 & 11276 & 10.23\% \\
         NL-50 & 50\% & 5734 & 1557 & 1.74\% \\
    \hline
    \end{tabular}
    }
\end{table}

We trained the models on the dataset in which the target image has at least $50\%$ of rain in the pixels (NL-50). This should set the focus of the trained networks on instances of rain. Something similar was done by the authors of \cite{xingjian2015convolutional} who select the top 97 rainy days of their dataset of three years for training. 

Furthermore, this enables the use of the dataset with at least $20\%$ of rain (NL-20) as an additional performance indicator. 
More precisely, we can use it as an indicator for the generalizability of the models. The trained models have not seen a single precipitation map of this test dataset. Furthermore, the models may have been biased towards predicting more rain due to the nature of predominantly rainy precipitation maps. Therefore it is possible that the performance of the models on this test set is worse than the one that closely resembles the data they are trained on.

\subsection{Cloud Cover Dataset}
\begin{figure*} 
    \centering
    \includegraphics[width=1.0\linewidth]{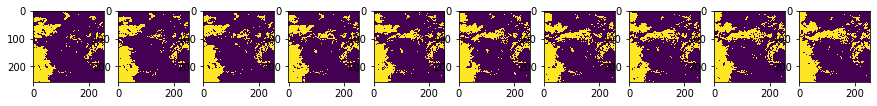}
    \caption{An example of cloud cover data sample.}
    \label{fig:cloud_cover_example}
\end{figure*}

Here, the cloud cover data introduced in \cite{berthomier2020cloudCover} is used as the second dataset to compare the proposed models. It contains binary images containing cloud position on each pixel with 1 indicating the cloud coverage and 0 indicating no cloud on the pixel. All images have size of 256x256 pixels and include the spatial area of France. Furthermore, each data sample consist of ten binary images: four images as an input and six images as the ground truth output. The images in each data sample are spaced by 15 minutes, which results in the time span of 1 hour for the input and 1 hour and 30 minutes for the output, totalling 2 hours and 30 minutes per sample. More details about this dataset can be found in \cite{berthomier2020cloudCover}. Exemplary sample of the discussed dataset is presented in Fig.~\ref{fig:cloud_cover_example}. 

As the cloud cover dataset includes images with binary values of 0 and 1 only, we do not apply any data normalization or image cropping. The cloud cover dataset is used for the training, evaluation and testing in its original form. The task of the network is to predict the probability of the presence of a cloud on each pixel.

\section{Results and Discussion}
\label{sec:discussion}

Following training of the four discussed models, we selected for each model the one with the lowest validation loss from their training run. These best performing models were then used to calculate several metrics, introduced in section \ref{sec:methods}, on the test set. The models were trained, evaluated and tested on the Precipitation map dataset as well as the cloud cover dataset separately.

\subsection{Evaluation on precipitation map dataset}

The results obtained on the Precipitation map dataset (NL-50) are tabulated in Table~\ref{tab:metric}. Note that the MSE is calculated after denormalizing the model predictions to the original rain intensities (mm/5min). Additionally, we calculated MSE values divided by the average pixel value of the two different datasets. The result is a normalized MSE (NMSE) with which we can have a fair comparison of error values between the different datasets.

\begin{table*}[]
    \centering
    \caption{MSE, NMSE and scores on rainfall bigger than $0.5mm/h$ indicating rain or no rain on the NL-50 dataset. Best result for that score is in bold. A $\uparrow$ indicates that higher values for that score are good whereas a $\downarrow$ indicates that lower scores are better.}
    \label{tab:metric}
    \resizebox{\linewidth}{!}{%
    \begin{tabular}{|c|c|c|c|c|c|c|c|c|c|c|}
        \hline
         Model & MSE $\downarrow$ & NMSE $\downarrow$ & Accuracy $\uparrow$ & Precision $\uparrow$ & Recall $\uparrow$ & F1 $\uparrow$ & CSI $\uparrow$ & FAR $\downarrow$ & HSS $\uparrow$ & Model size \\
         \hline
         Persistence (baseline) & 0.0248 & 847.67 & 0.756 & 0.678 & 0.643 & 0.660 & 0.493 & 0.320 & 0.235 & - \\
         UNet & \textbf{0.0122} & 416.38 & \textbf{0.836} & \textbf{0.740} & 0.855 & \textbf{0.794} & \textbf{0.658} & \textbf{0.259} & \textbf{0.329} & 1x  \\
         UNet with CBAM & 0.0171 & 584.46 & 0.820  & 0.707 & \textbf{0.871} & 0.780 & 0.640 & 0.293 & 0.315 & 1.01x \\
         UNet with DSC & 0.0127 & 435.86 & 0.812 & 0.700 & 0.856 & 0.770 & 0.626 & 0.300 & 0.306 & 0.23x \\
         SmaAt-UNet  & \textbf{0.0122} & \textbf{416.10} & 0.829 & 0.730 & 0.850 & 0.786 &  0.647 & 0.270 & 0.322 & 0.24x \\
         \hline
    \end{tabular}
    }
\end{table*}

The obtained results show that on the Precipitation map dataset, the common persistence baseline is outperformed by every model we tested by a large margin. This is noteworthy because, as mentioned before, it can be difficult to outperform this baseline in nowcasting due to the small time changes between the input and target.

We found that adding the proposed two modifications, i.e. DSCs and CBAMs, to the UNet architecture altered the models performance in comparison to the original UNet implementation on the Precipitation map dataset. On the one hand, implementing each modification alone slightly decreased the performance. On the other hand, however, our proposed model, SmaAt-Unet which incorporates both modifications into plain UNet, resulted in a better performance than UNet combined with each of the modifications alone. 
It should be noted that equipping UNet with only CBAMs, resulted in the highest MSE on the Precipitation map dataset with $0.0171$. 
Concerning our second modification, i.e. substituting the regular convolutions with DSCs, the results are more mixed. On the one hand, performance of the UNet with DSCs is worse than the original UNet model ($0.0127$ and $0.0122$, respectively). However, it still performs better than the UNet model with CBAMs. On the other hand, it is important to note that substituting regular convolutions by DSCs reduced the network's model size to a quarter of the original UNet. 

\begin{figure*}
    \centering
    \includegraphics[width=.7\linewidth]{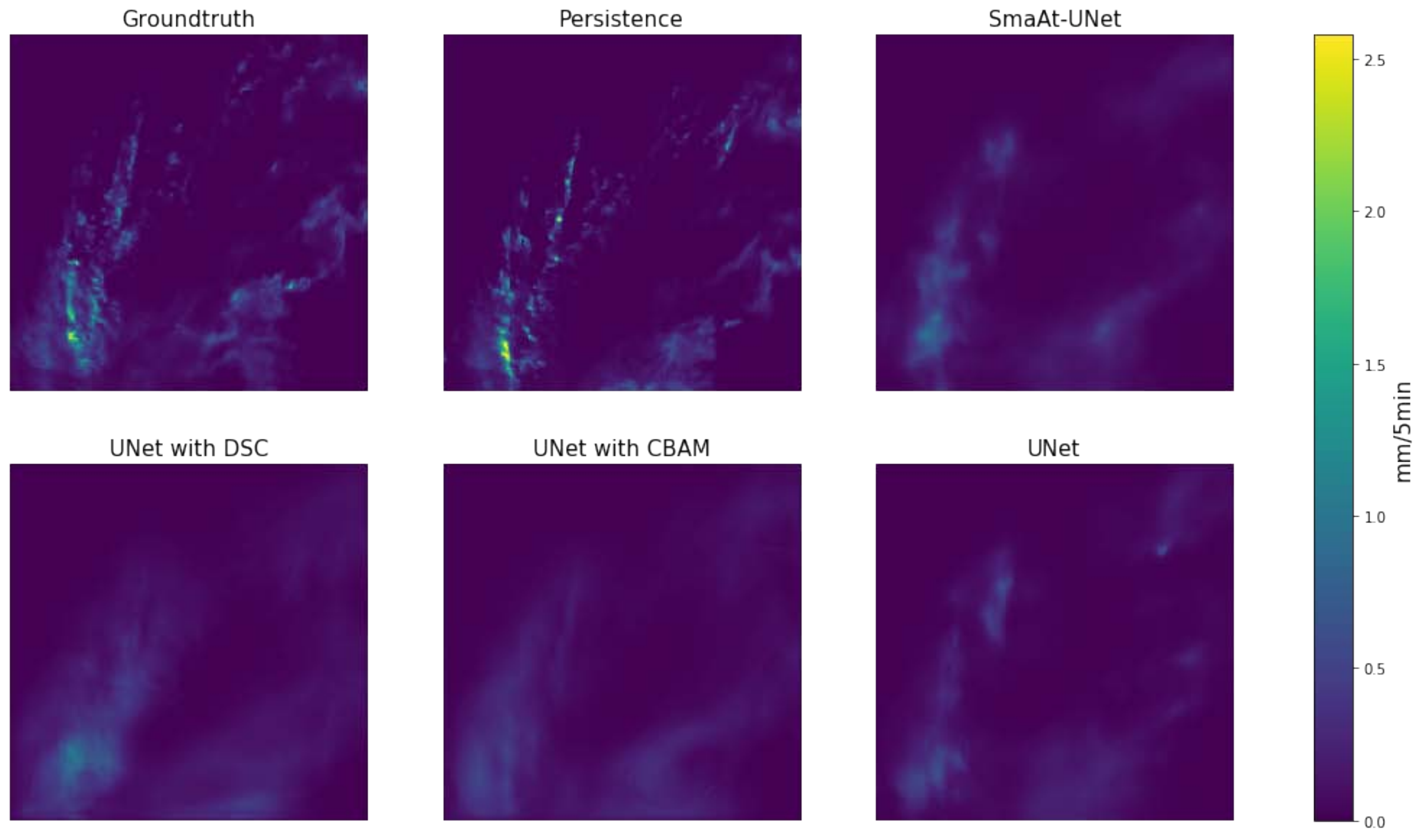}
    \caption{An example of precipitation nowcasting using the examined models.}
    \label{fig:precip_example}
\end{figure*}

Figure~\ref{fig:precip_example} shows an example of the models output for a precipitation nowcast on the Precipitation map dataset. In contrast to the ground truth image (top left panel) the predicted precipitation maps of all models are quite blurry. One reason for this is the use of MSE as guiding loss which is biased towards blurry images \cite{denton2018stochastic}. The bias towards blurriness is due to the fact that, given the many possibilities for future frames based on the input sequence, the model is trying to keep the error low by predicting a value that is closest to all possible outcomes \cite{mathieu2015deep}. Or, as Babaeizadeh~et al put it, "the models trained with a mean squared error loss function generate the expected value of all the possibilities for each pixel independently, which is potentially inherently blurry" \cite{babaeizadeh2017stochastic}.


Furthermore, one can see in Figure~\ref{fig:precip_example} that SmaAt-Unet is able to capture the development of intense rain clusters (lower left corner) better than the other models. UNet with DSCs predicts a spread that is too big on the horizontal elongation. UNet with CBAM does this better, but predicts values that are too conservative. UNet produces a similar output than SmaAt-UNet, but does not predict well enough the vertical spread of the precipitation of the left rain cluster.

Moreover, we have calculated several metrics of the performance of our models on the Precipitation map dataset. The obtained scores are also tabulated in Table~\ref{tab:metric}.
This table shows that while the original UNet implementation performs best in most scores, our SmaAt-UNet performs second best in six out of the seven scores. Thus, the SmaAt-Unet is able to approximate UNet's performance even though it only has 1/4 of its parameters. This facilitates research labs and individuals that do not possess a lot of computing power to also work on these computationally intensive calculations. This in turn can lead to a more rapid advancement in the development of radar-based short term rainfall prediction.

In order to test the generalizability of the models we use the other subset of our dataset that was described in section~\ref{sec:experiments}, i.e. NL-20. The MSE, NMSE and scores for this test set are given in Table~\ref{tab:metric_20}.
\begin{table*}[]
    \centering
    \caption{MSE and scores of the models on the test set from the NL-20 dataset. Calculated scores on rainfall bigger than $0.5mm/h$ indicating rain or no rain. Best result for that score is in bold. A $\uparrow$ indicates that higher values for that score are good whereas a $\downarrow$ indicates that lower scores are better.}
    \label{tab:metric_20}
    \resizebox{\linewidth}{!}{%
    \begin{tabular}{|c|c|c|c|c|c|c|c|c|c|c|}
        \hline
         Model & MSE $\downarrow$ & NMSE $\downarrow$ & Accuracy $\uparrow$ & Precision $\uparrow$ & Recall $\uparrow$ & F1 $\uparrow$ & CSI $\uparrow$ & FAR $\downarrow$ & HSS $\uparrow$ & Model size\\
         \hline
         Persistence (baseline) & 0.0227 & 1413.45 & 0.827 & 0.559 & 0.543 & 0.551 & 0.380 & 0.441 & 0.221 & -\\
         UNet               & \textbf{0.0111} & \textbf{691.48} & \textbf{0.880} & \textbf{0.666} & 0.782 & \textbf{0.719} & \textbf{0.562} & \textbf{0.334} & \textbf{0.321} & 1x\\
         UNet with CBAM     & 0.0147 & 913.40 & 0.860 & 0.607 & \textbf{0.812} & 0.695 & 0.532 & 0.393 & 0.303 & 1.01x \\
         UNet with DSC      & 0.0115 & 714.93 & 0.857 & 0.605 & 0.779 & 0.681 & 0.516 & 0.395 & 0.295 & 0.23x \\
         SmaAt-UNet         & \textbf{0.0111} & 692.08 & 0.867 & 0.626 & 0.801 & 0.703 & 0.542 & 0.374 & 0.309 & 0.24s \\
         \hline
    \end{tabular}
    }
\end{table*}
As can be seen in this table, the results are similar to the ones in Table~\ref{tab:metric}. Specifically, when ranking the models we can see that the original UNet implementation performs best in almost all metrics and our SmaAt-UNet comes in as close second in almost all metrics as well. This means that although the models have not seen many inputs with little rain, UNet and SmaAt-UNet are able to extrapolate best from the limited data that was available to them at training time. An explanation for the lower MSE in this dataset is that more values of the precipitation maps are close to zero (due to little rain) and therefore do not increase the overall MSE by a large margin if the model also predicts small values. Therefore, using NMSE for comparison is a better metric as it takes the pixel value distribution into account. Here, UNet is slightly better than SmaAt-UNet, but both their performance is way better than the other compared models. 

\begin{figure*}
    \centering
    \includegraphics[width=.65\linewidth]{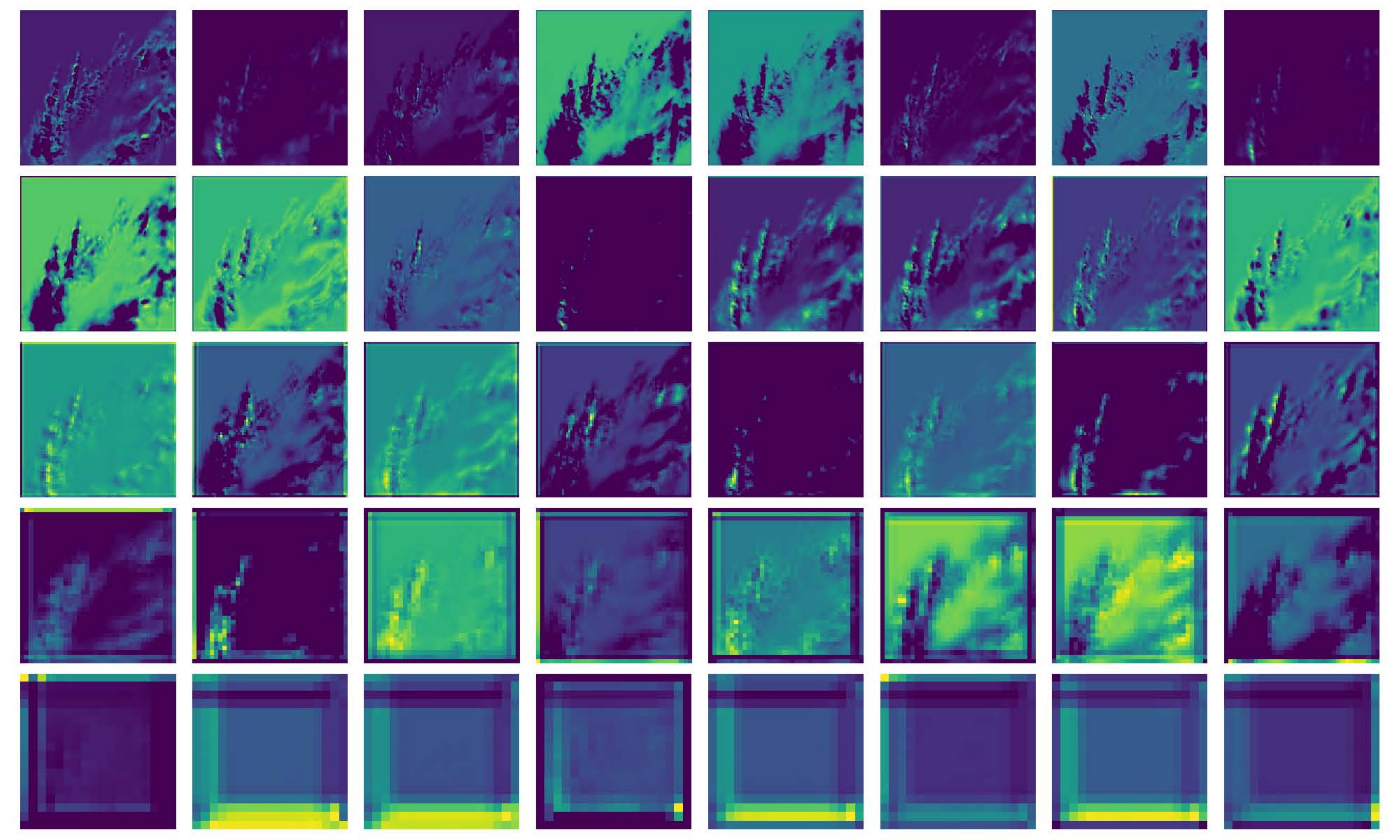}
    \caption{Example of 8 feature maps from each attention layer given an input sample. The same input sequence as was used in Fig.~\ref{fig:precip_example}. Top row to bottom row show examples of the five different attention layers. It is clear to see that the resolution of the images gets lower with each layer. In addition, it can also be seen that feature maps focus on different parts of the input.}
    \label{fig:attention_outputs}
\end{figure*}

Figure~\ref{fig:attention_outputs}, depicts example feature maps from the attention part of the encoder modules on the Precipitation map dataset. This figure illustrates that the network's attention maps learn to focus on particular parts of the input sequence, demonstrating the learning effect of the attention mechanism. The rows depict the different stages of the encoders which can be seen by a decrease in resolution in each row. Furthermore, it can be seen that the attention feature maps focus on different characteristics of the input. For example, in the first row, some feature maps focus on a rain cluster in the lower left corner (maps 2 and 8) while others focus on the parts with little to no rain (maps 4, 5 and 7). The bottom row shows feature maps from the last encoder stage of the SmaAt-UNet which have a resolution of $18 \times 18$. As the images in the bottom row illustrate, this low resolution leads the network to identify coarse patterns such as the rain cluster at the bottom of the maps (maps 2, 3, 5 and 7).

\subsection{Evaluation on cloud Cover dataset}
We also train the four discussed UNet-based models on the cloud cover dataset. The results of several metrics performance are tabulated in Table~\ref{tab:metric_cloud}. The cloud cover dataset contains samples with binary values and thus we do not calculate NMSE error here. Furthermore, the dataset format is set as four input images and six output images per sample.

\begin{table*}[]
    \centering
    \caption{MSE and scores on the cloud cover dataset \cite{berthomier2020cloudCover}. Best result for that score is in bold. A $\uparrow$ indicates that higher values for that score are good whereas a $\downarrow$ indicates that lower scores are better.}
    \label{tab:metric_cloud}
    \resizebox{\linewidth}{!}{%
    \begin{tabular}{|c|c|c|c|c|c|c|c|c|c|}
        \hline
         Model & MSE $\downarrow$ & Accuracy $\uparrow$ & Precision $\uparrow$ & Recall $\uparrow$ & F1 $\uparrow$ & CSI $\uparrow$ & FAR $\downarrow$ & HSS $\uparrow$ & Model size \\
         \hline
         UNet & 0.0785 & 0.890 & 0.895 & 0.919 & \textbf{0.907} & 0.829 & 0.105 & 0.386 & 1x  \\
         UNet with CBAM & \textbf{0.0775} & \textbf{0.891}  & \textbf{0.902} & 0.913 & \textbf{0.907} & \textbf{0.831} & \textbf{0.098} & \textbf{0.388} & 1.01x \\
         UNet with DSC & 0.0793 & 0.889 & \textbf{0.902} & 0.908 & 0.905 & 0.827 & \textbf{0.098} & 0.386 & 0.23x \\
         SmaAt-UNet  & 0.0794 & 0.889 & 0.892 & \textbf{0.921} & 0.906 &  0.829 & 0.108 & 0.385 & 0.24x \\
         \hline
    \end{tabular}
    }
\end{table*}

From Table~\ref{tab:metric_cloud}, one can observe that the lowest MSE score belongs to UNet with CBAM. However, the results are comparable for all models. The difference between the highest and the lowest MSE obtained on the cloud cover dataset for all the four models examined is $0.0019$. The small difference is particularly worth mentioning for SmaAt-UNet and Unet with DSC, which contain roughly 1/4 of the parameters compared to UNet or UNet with CBAM.


For most of the reported metrics in Table~\ref{tab:metric_cloud}, UNet with CBAM reaches the best scores. In two cases UNet with CBAM is comparable to another model, i.e. with UNet for F1 score and with UNet with DSC for FAR score. For this dataset SmaAt-Unet yields the best Recall score. Similarly as for MSE score, the other reported scores are also comparable and the differences are minor. It shows that on cloud cover dataset, on which the model task is to predict the cloud probabilities between 0 and 1, the proposed combination of UNet with convolutional block attention modules reaches the best scores in most of the cases. Nevertheless, the proposed SmaAt-UNet reaches very similar performance with approximately $1/4$ parameters of original UNet and UNet with CBAM.

It should be noted that in case of the cloud cover dataset, the differences between the scores are smaller compared to those of the precipitation map dataset, because the data values of the cloud cover dataset are binary. The evaluated models predict values between 0 and 1 for the cloud cover data (presence or absence of the cloud) which results in smaller differences compared to the evaluation on the precipitation map dataset (see Table~\ref{tab:metric} and Table~\ref{tab:metric_20}).

\section{Conclusion}
\label{sec:conclusion}

In this paper we proposed SmaAt-UNet which is a smaller and attentive version of a UNet architecture. It has been shown that it performs on par to similar architectures that are way bigger than itself on a precipitation nowcasting task. 
The development of small and efficient neural networks, such as SmaAt-UNet, enables their application in smartphones. For instance, creating an application with multiple trained SmaAt-Unets with different forecasting times allows precipitation forecasting with the latest available data at the users request. Furthermore, creating energy efficient architectures, such as SmaAt-UNet, reduces the carbon footprint. 
Being mindful of the resources that are required for training a neural network is a crucial step towards sustainable machine learning practices.


\bibliography{main}

\end{document}